\documentclass[10pt,twocolumn,letterpaper]{article}

\usepackage[pagenumbers]{wacv}      
\usepackage{graphicx}
\usepackage{amsmath}
\usepackage{amssymb}
\usepackage{booktabs}

\usepackage{bm}
\usepackage{diagbox}
\usepackage{multirow}
\newcommand{\shortsection}[1]{\noindent\textbf{#1.}}
\newtheorem{theorem}{Theorem}

\usepackage{color}
\newcommand{\ifcomments}{\iftrue}
\definecolor{ForestGreen}{cmyk}{0.864, 0.0, 0.429, 0.396}

\usepackage[pagebackref,breaklinks,colorlinks]{hyperref}

\usepackage[capitalize]{cleveref}
\crefname{section}{Sec.}{Secs.}
\Crefname{section}{Section}{Sections}
\Crefname{table}{Table}{Tables}
\crefname{table}{Tab.}{Tabs.}

\def\wacvPaperID{348} 
\def\confName{WACV}
\def\confYear{2025}

\begin{document}

\title{DiffPAD: Denoising Diffusion-based Adversarial Patch Decontamination}

\author{Jia Fu$^{1,2}$\quad Xiao Zhang$^3$\quad Sepideh Pashami$^{1,4}$\quad Fatemeh Rahimian$^1$\quad Anders Holst$^{1,2}$\quad\\
$^1$RISE Research Institutes of Sweden\quad $^2$KTH Royal Institute of Technology\\
$^3$CISPA Helmholtz Center for Information Security\quad $^4$Halmstad University\\
{\tt\small \{jia.fu, sepideh.pashami, fatemeh.rahimian, anders.holst\}@ri.se\quad xiao.zhang@cispa.de}
}
\maketitle

\begin{abstract}
   In the ever-evolving adversarial machine learning landscape, developing effective defenses against patch attacks has become a critical challenge, necessitating reliable solutions to safeguard real-world AI systems. Although diffusion models have shown remarkable capacity in image synthesis and have been recently utilized to counter $\ell_p$-norm bounded attacks, their potential in mitigating localized patch attacks remains largely underexplored. In this work, we propose \textbf{DiffPAD}, a novel framework that harnesses the power of diffusion models for adversarial patch decontamination. DiffPAD first performs super-resolution restoration on downsampled input images, then adopts binarization, dynamic thresholding scheme and sliding window for effective localization of adversarial patches. Such a design is inspired by the theoretically derived correlation between patch size and diffusion restoration error that is generalized across diverse patch attack scenarios. Finally, DiffPAD applies inpainting techniques to the original input images with the estimated patch region being masked. By integrating closed-form solutions for super-resolution restoration and image inpainting into the conditional reverse sampling process of a pre-trained diffusion model, DiffPAD obviates the need for text guidance or fine-tuning. Through comprehensive experiments, we demonstrate that DiffPAD not only achieves state-of-the-art adversarial robustness against patch attacks but also excels in recovering naturalistic images without patch remnants. The source code is available at \url{https://github.com/JasonFu1998/DiffPAD}.
\end{abstract}


\section{Introduction}
\label{sec:introduction}

Despite achieving remarkable success in a wide range of machine learning applications, deep neural networks (DNNs) are extremely susceptible to adversarial examples~\cite{szegedy2014intriguing}, normal inputs crafted with small perturbations that are devised to induce model errors. The discovery of adversarial examples raises serious concerns about the robustness of DNNs, particularly for security-sensitive domains.
Existing works primarily focus on $\ell_p$-norm bounded perturbations~\cite{goodfellow2015explaining, 7958570, madry2018towards}, a specific type of global attacks, where the adversary is allowed to manipulate all the pixels within the entire image. In contrast, adversarial patch attacks~\cite{brown2017adversarial, karmon2018lavan, xiang2021gdpa} restrict the total number of pixels that can be modified, which typically occupy a small localized region within the image. Since adversarial patches can be easily attached to physical-world objects~\cite{wei2022adversarial}, they pose more realistic threats to security-critical DNN systems, ranging from surveillance cameras~\cite{xu2020adversarial} to autonomous vehicles~\cite{eykholt2018robust}. Therefore, it is crucial to develop effective defenses that are robust to adversarial patch attacks.


Numerous defenses have been proposed to enhance the robustness of DNNs, such as adversarial purification~\cite{samangouei2018defensegan}, adversarial training~\cite{madry2018towards} and certified defenses~\cite{wong2018provable}. In particular, adversarial purification~\cite{song2018pixeldefend,shi2020online,hill2020stochastic} leverages the power of generative models to remove adversarial noise. Compared with adversarial training and certified defenses, adversarial purification has the potential to protect the target model without the need for adaptive retraining. Witnessing the exceptional capability of diffusion models for image synthesis tasks~\cite{ho2020denoising,dhariwal2021diffusion}, recent works utilize diffusion models for adversarial defenses~\cite{nie2022DiffPure,carlini2022certified,wang2023better}, which achieve state-of-the-art performance in building robust models.
However, all these methods are designed to defend against global attacks. When such defenses encounter localized patch attacks, their performance will drop to different extents, failing to fulfill the security requirements of real-world applications. It remains unclear how adversarial purification and diffusion models can help mitigate adversarial patch attacks.
In order to defend against localized patch attacks, more specialized strategies, such as adversarial patch detection and segmentation~\cite{liu2022segment,tarchoun2023jedi}, have been developed to isolate and neutralize adversarial patches before processing images through the model. These methods leverage the advancements in digital image processing to detect anomalies that are indicative of patch tampering. Nevertheless, these methods often struggle to reconstruct the original images with high fidelity and are not successful in defending against adaptive attacks that can avoid gradient obfuscations~\cite{athalye2018obfuscated}.

To address the limitations of previous defense strategies and exploit the full potential of diffusion models in defending against patch attacks, we propose \textbf{DiffPAD}, a \textbf{Diff}usion-based framework for adversarial \textbf{PA}tch \textbf{D}econtamination. Figure~\ref{fig:teaser} depicts the workflow of DiffPAD. By decomposing the defense task into patch localization and inpainting, we address these two sub-problems via a diffusion model's conditional reverse sampling process. Such a conditional process incorporates the visual information of the clean region to retain label semantics integrally. We also show evidence that justifies the usefulness of conditional diffusion models in mitigating the distribution deviation caused by adversarial patch variations, and achieve effective patch localization through a single diffusion sampling process, where the discrepancy in adversarial region is accentuated by resolution degradation and restoration techniques. Comprehensive experiments demonstrate the substantial advancement of DiffPAD in adversarial patch defense over existing methods, offering a robust and scalable solution that aligns with real-world application requirements. Our main contributions are further summarized as follows:


\begin{itemize}
    \item We integrate the closed-form solution of image super-resolution into the reverse sampling process of a pre-trained diffusion model for patch localization, eliminating the need for fine-tuning and multiple reverse generations. Patch decontamination is then accomplished using the same diffusion process, but by switching the closed-form solution to inpainting.
    \item We prove a linear correlation between the patch size and an upper bound of diffusion restoration error. Such a relationship is empirically verified across different classification models and patch attacks with varying patch sizes and random positions, which facilitates the efficiency and accuracy of patch detection.
    \item Through comprehensive experiments on image classification and facial recognition tasks, we demonstrate that DiffPAD achieves state-of-the-art adversarial robustness against both adaptive and non-adaptive patch attacks, and is capable of completely removing patch remnants and generating naturalistic images.
\end{itemize}

\section{Related work}
\label{sec:related work}

This section introduces the most related works to ours. Other discussions are provided in supplementary materials.

\vspace{0.1cm}

\shortsection{Defenses against adversarial patches} 
To enhance the model robustness against adversarial attacks, various defense strategies have been proposed. 
Initial attempts focused on simple preprocessing-based defenses, such as JPG compression~\cite{dziugaite2016study}, thermometer encoding~\cite{buckman2018thermometer}, defensive distillation~\cite{papernot2016distillation}. However, these methods have been shown ineffective against adaptive attacks~\cite{athalye2018obfuscated}. Adversarial training~\cite{madry2018towards}, which optimizes the neural network parameters by incorporating adversarially generated inputs in training, is by far the most popular. Nevertheless, adversarial training suffers from high computational costs, due to the iterative steps required for generating strong adversarial examples. Certified defenses~\cite{wong2018provable} have also been developed, 
but they cannot achieve a similar level of robustness to adversarial training. When adversarial training and certified defenses are applied to defend against adversarial patches~\cite{rao2020adversarial,chiang2020certified}, the learned model is usually only effective to the specific attacks employed in training but shows inferior generalization performance to unseen patch attacks.
To achieve comparable robustness against adversarial patches, more specialized patch detection schemes have been developed to localize and purify adversarial patches. For instance, Liu~\textit{et al.}~\cite{liu2022segment} trained a patch segmenter to generate pixel-level masks for adversarial patches, while Tarchoun~\textit{et al.}~\cite{tarchoun2023jedi} 
identified patch localization based on the property that the entropy of adversarial patches is higher compared with other regions. Our work falls into the line of patch detection-based defenses, but aims to leverage the power of diffusion models to achieve better defense performance.




\vspace{0.1cm}

\shortsection{Adversarial purification} Adversarial purification~\cite{samangouei2018defensegan,song2018pixeldefend,shi2020online,nie2022DiffPure} refers to a special family of preprocessing-based defenses that makes use of generative models.
For example, DiffPure~\cite{nie2022DiffPure} gradually injects Gaussian noise in the forward diffusion steps followed by denoising during the reverse generation phase, where the adversarial noise is purified along the process. Such state-of-the-art generative models offer a promising avenue for mitigating adversarial examples~\cite{wang2022guided, xiao2023densepure}. Nevertheless, adversarial purification frameworks are typically designed for purify $\ell_p$-norm bounded perturbations, which may fall short against the discrete and localized nature of adversarial patches. So far, we have only identified a single existing method, DIFFender~\cite{kang2023diffender}, that utilizes diffusion models to defend against patch attacks. DIFFender adopts a text-guided diffusion model to localize the adversarial patch and then reconstruct the original image. Unfortunately, DIFFender not only requires expensive multiple reverse diffusion processes for effective patch localization, but also relies on heuristic manually-designed prompts or complex prompt tuning steps, hindering automation in practical use cases.



\vspace{-0.1cm}
\section{Preliminaries on diffusion models}
\label{sec:preliminary}

A diffusion model consists of forward and reverse diffusion processes. The forward process progressively degrades the underlying distribution $p_0$ towards a noise distribution by adding Gaussian noise, which can be characterized by a stochastic differential equation (SDE)~\cite{song2021scorebased}:
\begin{equation}
\label{e1}
\mathrm{d} \bm{x}=\bm{f}(\bm{x}, t) \mathrm{d} t+g(t) \mathrm{d} \bm{w},
\end{equation}
where $\bm{x}_t\in\mathbb{R}^d$ follows $p_t$ denoting the distribution at time step $t$, $\bm{w}$ denotes the standard Wiener process (a.k.a. Brownian motion), and $\bm{f}$: $\mathbb{R}^d \times \mathbb{R}\rightarrow \mathbb{R}^d$ and $g$: $\mathbb{R} \rightarrow \mathbb{R}$ represent the drift and diffusion coefficients, respectively. Given the distribution $p_t$, the reverse diffusion process with respect to Equation \ref{e1} can be formulated as:
\begin{equation}
\label{e2}
\mathrm{d} \bm{x}=\left[\bm{f}(\bm{x}, t)-g(t)^2 \nabla_{\bm{x}} \log p_t(\bm{x})\right] \mathrm{d} t+g(t) \mathrm{d} \bm{w}.
\end{equation}
For generation tasks where the condition $\bm{y}$ is specified, the objective is to sample data from $p(\bm{x}|\bm{y})$. By applying Bayes' theorem, the conditional process with respect to Equation \ref{e2} can be written as:
\begin{align}
\label{e3}
\nonumber\mathrm{d} \bm{x}=\big[\bm{f}(\bm{x}, t)-g(t)^2 \nabla_{\bm{x}}\big(\log p_t(\bm{x}) &+ \log p_t(\bm{y} | \bm{x})\big)\big] \mathrm{d} t \\ 
 &\quad+ g(t) \mathrm{d} \bm{w}.
\end{align}

\vspace{0.1cm}

\shortsection{Denoising diffusion probabilistic models (DDPMs)}
DDPM~\cite{ho2020denoising} is a milestone in diffusion models, which offers unparalleled stability and quality for generative tasks. Specifically, DDPM models the generative process using a Markov chain $\bm{x}_T \rightarrow \bm{x}_{T-1} \rightarrow \ldots \rightarrow \bm{x}_0$, where the joint distribution is defined as:
\begin{equation}
\label{e4}
p_\theta\left(\bm{x}_{0: T}\right)=p\left(\bm{x}_T\right) \prod_{t=1}^T p_\theta\left(\bm{x}_{t-1} | \bm{x}_t\right).
\end{equation}
DDPM sets $\bm{f}(\bm{x}, t) = -\frac{1}{2} \beta_t \bm{x}$ and $g(t) = \sqrt{\beta_t}$ and derives the discrete-time diffusion processes. According to the statistical properties of Gaussian distribution, DDPM samples $\bm{x}_t$ from $\bm{x}_0$ using the following closed-form solution:
\begin{align}
\label{e6}
    \bm{x}_t = \sqrt{\bar{\alpha}_t}\bm{x}_0 + \sqrt{1 - \bar{\alpha}_t}\bm{\epsilon},
\end{align}
where $\bar{\alpha}_t = \Pi_{s=1}^t (1-\beta_s)$ and $\bm\epsilon\sim\mathcal{N}(0,\mathbf{I})$. For the reverse sampling, DDPM estimates $\bm{x}_0$ based on the following approximation:
\begin{equation}
\label{e5}
\bm{x}_0 \approx \hat{\bm{x}}_0 = \frac{1}{\sqrt{\bar{\alpha}_t}} \left(\bm{x}_t-\sqrt{1-\bar{\alpha}_t}\boldsymbol{\epsilon}_\theta\left(\bm{x}_t\right)\right),
\end{equation}
where $\boldsymbol{\epsilon}_\theta$ denotes the neural network designed to predict the total noise between $\bm{x}_t$ and $\bm{x}_0$ based on Equation \ref{e6}. 

\vspace{0.1cm}

\shortsection{Denoising diffusion restoration models}
Given the strong sample quality of DDPMs in image synthesis, DDPMs have been adapted for conditional use in image restoration tasks~\cite{choi2021ilvr, kawar2022denoising, zhu2023denoising}. Kawar \textit{et al.}~\cite{kawar2022denoising} first introduced the term ``DDRM'', where the condition $\bm{y}$ is a degraded image of $\bm{x}$ and the distribution of DDRM is defined as:
\begin{equation}
\label{e7}
p_\theta\left(\bm{x}_{0: T}| \bm{y}\right)=p\left(\bm{x}_T| \bm{y}\right) \prod_{t=1}^T p_\theta\left(\bm{x}_{t-1} | \bm{x}_t, \bm{y}\right).
\end{equation}
In the following discussions, we refer to the family of methods that conditionally utilize DDPMs for image restoration, including the first work~\cite{kawar2022denoising}, as DDRMs. During the reverse generation process, DDRMs replace $\hat{\bm{x}}_0$ by its $\bm{y}$-conditioned counterpart $\Tilde{\bm{x}}_0$. Variations among different DDRM models primarily arise in the computation of $\Tilde{\bm{x}}_0$ from $\hat{\bm{x}}_0$ and $\bm{y}$. For instance, the vanilla DDRM proposed a singular value decomposition (SVD) based approach, assuming a linear degradation function to compute $\Tilde{\bm{x}}_0$ by pseudo-inverse, whereas DiffPIR~\cite{zhu2023denoising} calculated $\Tilde{\bm{x}}_0$ by solving a data proximal sub-problem.



\section{Proposed method: DiffPAD}
\label{sec:methodology}

\begin{figure*}
  \centering
  \begin{minipage}[c]{0.63\linewidth}
    \includegraphics[width=1\textwidth]{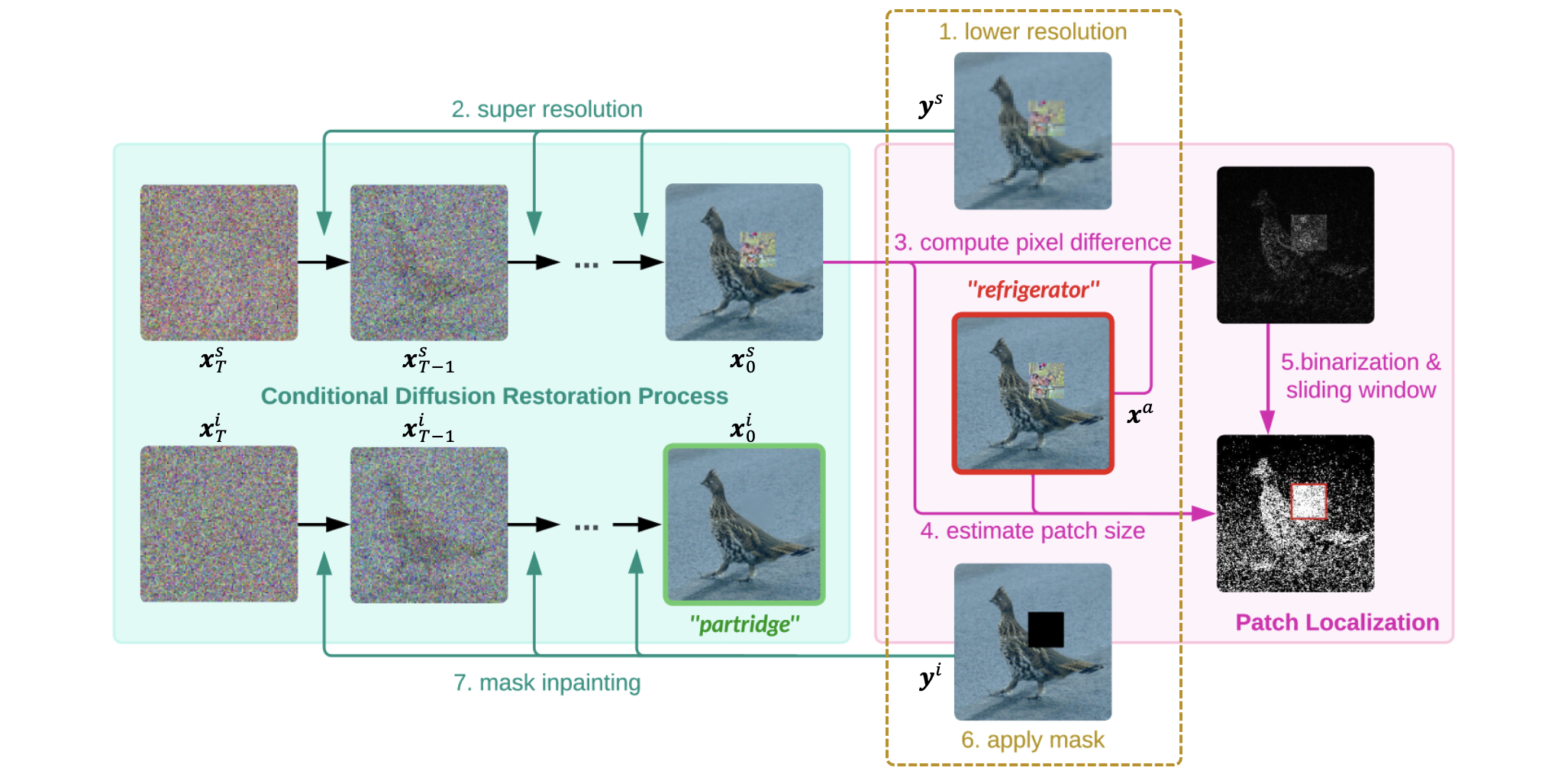}
    \caption{The overall pipeline of DiffPAD, which follows steps numbered from 1 to 7 in order. Text and blocks in turquoise, pink and yellow correspond to the conditional diffusion restoration module, patch localization module and image degradation operations, respectively. The input of DiffPAD is the adversarial patch contaminated image $\bm{x}^a$ (with red frame), and the output is the decontaminated image $\bm{x}_0^i$ (with green frame).}
    \label{fig:teaser}
  \end{minipage}
  \hfill
  \begin{minipage}[c]{0.35\linewidth}
    \includegraphics[width=1\linewidth]{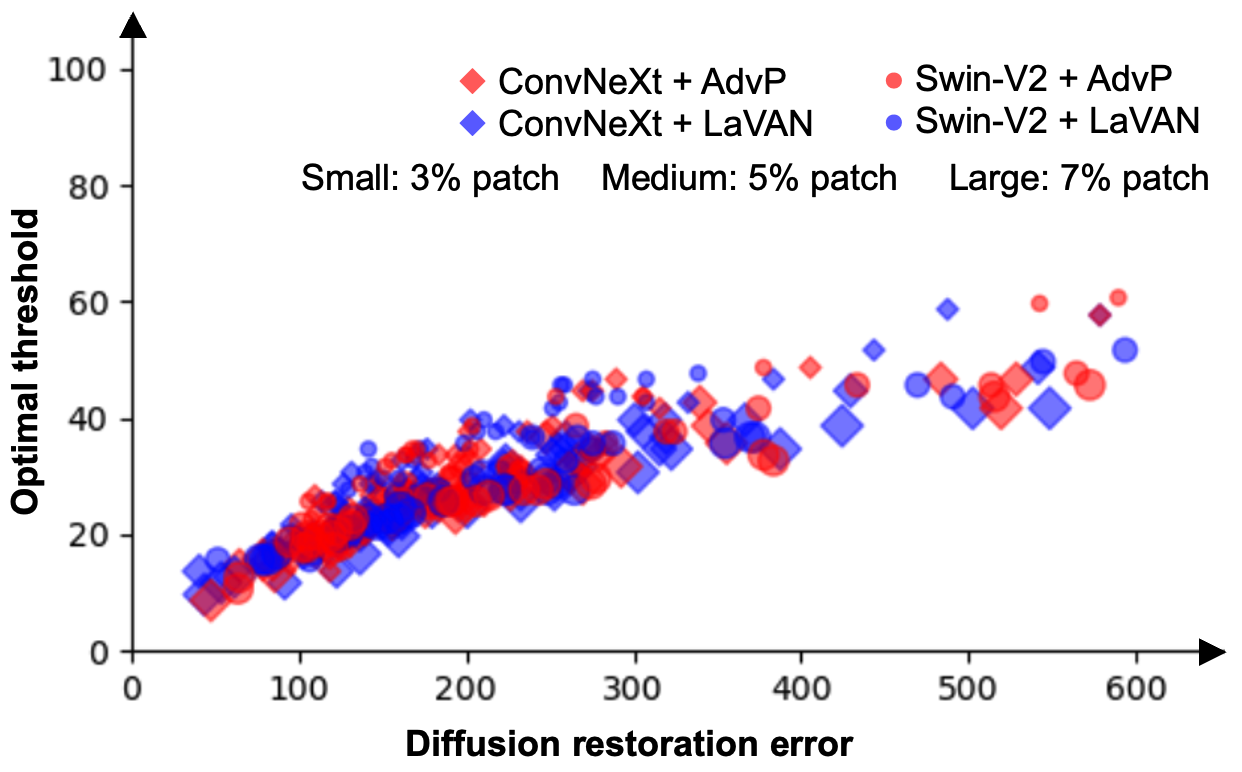}
    \caption{Illustration of the linear relationship between diffusion restoration errors and optimal thresholds for patch localization under various attacks. In particular, we vary the size of the adversarial patches generated by different attacks on various model architectures.}
    \label{fig:Linear}
  \end{minipage}
\end{figure*}


In this section, we first motivate and explain our design of DiffPAD that utilizes conditional diffusion models for accurate patch localization followed by patch restoration.
Mathematically, we work with the following definition of adversarial patches. Let $\bm{x}^c$ denote the clean image and $\pmb{\delta}$ be the adversarial perturbation. Then, the adversarial patch contaminated image can be defined as: 
\begin{equation}
\label{e8}
\bm{x}^a=(\mathbf{1}-\mathbf{A})\odot \bm{x}^c+\mathbf{A} \odot \pmb{\delta},
\end{equation}
where $\mathbf{A}\in\{0,1\}^d$ is the mask of the region outside the adversarial patch, and $\odot$ denotes the Hadamard product.
Unlike global attacks, adversarial patches modify only a small localized region of the clean image, concealing its original visual information.

\subsection{DDRMs for patch defenses} 
DiffPAD follows a stepwise pipeline of patch restoration after its localization, which utilizes DDRMs for defending against patch attacks (Figure \ref{fig:teaser}).
Equation \ref{e8} reveals that the label semantics of $\bm{x}^a$ comes from $(\mathbf{1}-\mathbf{A})\odot \bm{x}^c$, i.e., the clean region. The clean region itself serves as the optimal condition for guiding the diffusion process to keep the image semantics of the original clean image to the greatest extent. Although $\mathbf{A}$ is unknown for a given adversarial example, all information from $(\mathbf{1}-\mathbf{A})\odot \bm{x}^c$ is included in $\bm{x}^a$, suggesting an approach to construct the condition based on $\bm{x}^a$. If we denote $\mathcal{H}$ as an image degradation function and set $\bm{y}=\mathcal{H}(\bm{x}^a)$, the corresponding diffusion process naturally translates to DDRMs. A moderate image degradation function, such as image compression, typically preserves enough image semantics to allow high-quality restoration by DDRMs. Employing DDRMs as the foundation model of DiffPAD frees us from selecting a specific halt time step or introducing extra text prompts to keep label semantics during reverse diffusion sampling. Based on variational inference, the ELBO objective of DDRMs can be rewritten in the form of DDPMs objective, as shown in Theorem 3.2 in~\cite{kawar2022denoising}, which supports the feasibility of approximating the optimal solutions of our DDRM-based framework by pre-trained DDPM models without any fine-tuning. The Gaussian noise of DDPMs is in nature the discretization of the variance preserving SDE~\cite{song2021scorebased}, so does DDRMs. Therefore, the following property held in the forward process of DDPMs also holds for DDRMs:
\begin{equation}
\label{e9}
\frac{\partial D_{\mathrm{KL}}\left(p^c_t \| p^a_t\right)}{\partial t} \leq 0,
\end{equation}
where $D_{\mathrm{KL}}$ denotes the Kullback-Leibler divergence. Equation \ref{e9} implies that the injected Gaussian noise will disrupt the patterns of adversarial patches, gradually aligning the adversarial distribution $p^a$ with the clean distribution $p^c$ through the forward DDRMs process. 


\vspace{0.1cm}

\shortsection{Resolution degradation and restoration} We aim to localize the patch according to the property that the adversarial region exhibits more drastic changes compared to the clean region when $\bm{x}^a$ is compared with its diffusion-generated counterpart. However, as demonstrated in \cite{kang2023diffender}, solely relying on the stochasticity of the diffusion process to disrupt the patch pattern is inefficient. To address this challenge, we propose a \emph{resolution degradation-restoration mechanism} to amplify the diffusion restoration error in the patch region.
Such a design is motivated by the observation that image compression can enhance the model robustness against patch attacks, suggesting the high sensitivity of adversarial patches to resolution changes. In DiffPAD, we first employ bicubic down-sampling with a scaling factor $s$ on $\bm{x}^a$ to obtain $\bm{y}^s$, serving as the initial destruction to the adversarial patch distribution, then intensify this destruction through the randomness along with DDRMs super-resolution restoration, conditioned on $\bm{y}^s$ (Steps 1-2 in Figure \ref{fig:teaser}). This preparation allows DiffPAD to precisely localize the adversarial patch through a single diffusion generation, whereas DIFFender necessitates the generation of at least three samples per image to ensure robust patch localization. Different from the vanilla DDRM, which assumes that $\mathcal{H}$ is linear and uses an SVD solution to compute $\Tilde{\bm{x}}_0$. DiffPAD leverages a fast closed-form solution for efficient diffusion restoration. Denoting $\eta_t = \bar{\alpha}_t \sigma^2 / (1-\bar{\alpha}_t)$ where $\sigma$ represents the noise level associated with $\bm{y}$, we adopt a plug-and-play image resolution restoration function from~\cite{zhang2021plug} with a bicubic kernel $\mathbf{k}$:
\begin{equation}
\label{e10}
\Tilde{\bm{x}}_0=\mathcal{F}^{-1}\left(\frac{1}{\eta_t}\left(\mathbf{d}-\overline{\mathcal{F}}(\mathbf{k}) \odot_s \frac{(\mathcal{F}(\mathbf{k}) \mathbf{d}) \downarrow_s}{(\overline{\mathcal{F}}(\mathbf{k}) \mathcal{F}(\mathbf{k})) \downarrow_s+\eta_t}\right)\right),
\end{equation}
where $\mathbf{d}=\overline{\mathcal{F}}(\mathbf{k}) \mathcal{F}\left(\bm{y}^s \uparrow_{s}\right)+\eta_t \mathcal{F}\left(\hat{\bm{x}}_0\right)$. $\mathcal{F}$, $\mathcal{F}^{-1}$, and $\overline{\mathcal{F}}$ denote Fast Fourier Transform (FFT), its inverse, and conjugate, respectively. As for operators, $\uparrow_{s}$ is the standard $s$-fold up-sampler, $\downarrow_{s}$ is the down-sampler that averages $s\times s$ distinct blocks, and $\odot_{s}$ is element-wise multiplication for distinct block processing. 

\vspace{0.1cm}
\shortsection{Inpainting} By substituting Equation \ref{e10} with the closed-form solution for inpainting restoration, we can take advantage of the same pre-trained DDPMs used in super-resolution restoration. Once the patch is localized and masked, we adapt a plug-and-play color image demosaicing function from~\cite{zhang2021plug} for the image inpainting task:
\begin{equation}
\label{e11}
\Tilde{\bm{x}}_0=\frac{\mathbf{M} \odot \bm{y}^i+\eta_t \hat{\bm{x}}_0}{\mathbf{M}+\eta_t},
\end{equation}
where $\mathbf{M}\in\{0,1\}^d$ is a customized mask on $\bm{x}^a$ for acquisition of $\bm{y}^i$. Note the division operations in Equations \ref{e10} and \ref{e11} are element-wise.

\subsection{Adversarial patch localization}
This section explains the algorithm for estimating patch size and localizing its position. We adopt the most common setting for adversarial patches: crafting a single square-shaped patch of random size and position on a given clean image.
Typically, the size of adversarial patches should not be too large, as it will block the label semantics and render the image unrecognizable to humans. 
Inspired by Theorem 3.2 of \cite{nie2022DiffPure}, which proves an upper bound on the $\ell_2$ distance between a diffusion-purified $\ell_p$-norm bounded adversarial example and the corresponding clean image, we establish a similar result in the following theorem for patch attacks.

\begin{theorem}
\label{thm:connection}
Assume $\|\bm\epsilon_\theta\left(\bm{x}_t\right)\|\leq C_\epsilon\sqrt{1-\bar{\alpha}_t}$ and let $\gamma:=\int_0^{T} \beta_t \mathrm{d} t$.  With probability at least $1-\xi$, the $\ell_2$ distance between the diffusion-purified image $\hat{\bm{x}^a}$ with adversarial patch and the corresponding clean image $\bm{x}^c$ satisfies:
\begin{equation}
\|\hat{\bm{x}^a}-\bm{x}^c\| \leq \varepsilon\left|\mathbf{A}\right|+\gamma C_\epsilon+\sqrt{e^{\gamma}-1} \cdot C_\xi,
\end{equation}
where $\varepsilon$ is the $\ell_2$-norm bound of the patch, $C_\xi:=\sqrt{2 d+4 \sqrt{d \log \frac{1}{\xi}}+4 \log \frac{1}{\xi}}$, and $d$ is the input dimension.
\end{theorem}

The proof of Theorem 1 is provided in the supplementary materials.
Note that $\|\bm{x}^c-\bm{x}^a\| \leq\varepsilon\left|\mathbf{A}\right|$, then with the help of the triangle inequality, the $\ell_2$ distance between a patch-contaminated image before and after diffusion reconstruction can be upper bounded as:
\begin{equation}
\label{linear}
\|\hat{\bm{x}^a}-\bm{x}^a\| \leq 2\varepsilon\left|\mathbf{A}\right|+\gamma C_\epsilon+\sqrt{e^{\gamma}-1} \cdot C_\xi,
\end{equation}
where the bound is linearly correlated to the patch area $\left|\mathbf{A}\right|$. In practice, directly estimating the patch size based on $\|\hat{\bm{x}^a}-\bm{x}^a\|$ yields unsatisfactory results. The upper bound of $\ell_2$ distance primarily emphasizes the most significant differences arising from the restoration. In other words, the subtle variations caused by the intrinsic randomness of the diffusion model should be neglected. 

Denote $\ominus$ as the pixel-wise difference and $\bm{x}^\Delta := \hat{\bm{x}^a} \ominus \bm{x}^a$. As previously justified, the adversarial regions with nearly full-scale exhibit higher discrepancies in $\bm{x}^\Delta$, a result of our patch error amplification implemented during the diffusion restoration on resolution (Step 3 in Figure \ref{fig:teaser}). To isolate the pixels that contribute most to $\|\hat{\bm{x}^a}-\bm{x}^a\|$ and count their quantity to represent the patch area, we apply the binarization on $\bm{x}^\Delta$ with dynamic threshold $\tau$. Enlightened by Equation~\ref{linear}, we posit that $\tau$ more reasonably reflects the upper bound of $\|\hat{\bm{x}^a}-\bm{x}^a\|$. A higher $\tau$ indicates that the pixels filtered out by binarization are likely to have a higher diffusion restoration error, suggesting a higher probability of originating from the patch region, equivalently, a larger patch area in a given image. Therefore, we propose to model $\tau$ as being linearly correlated with the mean squared error (MSE) between $\hat{\bm{x}^a}$ and $\bm{x}^a$:
\begin{equation}
\label{eq:localization threshold}
\tau = \mu \cdot \frac{\|\hat{\bm{x}^a}-\bm{x}^a\|}{d}+\nu,
\end{equation}
where $\mu$ and $\nu$ are hyperparameters. Consequently, the estimated patch area will be $\Tilde{A}= \left| \mathrm{Binarize}(\bm{x}^\Delta, \tau)\right|$ (Step 4 in Figure \ref{fig:teaser}).
To illustrate the linear relation, we craft adversarial examples from the ImageNet dataset with varying patch areas ($3\%, 5\%, 7\%$ of full image size) and random positions using different attack mechanisms (i.e., AdvP and LaVAN) across diverse classifiers (i.e., ConvNeXt and Swin-V2). Each attack consists of 25 examples. We conduct a traversal search to identify the optimal threshold for binarization on  $\bm{x}^\Delta$, which provides the most accurate estimation of the original patch area for each example. By depicting the relation between the optimal thresholds and the corresponding MSE values, Figure \ref{fig:Linear} confirms their linear correlation and validates the effectiveness of our estimation method.

After obtaining the estimated patch size, DiffPAD uses a sliding window of the same size to scan the values of $\mathrm{Binarize}(\bm{x}^\Delta, \tau')$, where $\tau'$ is a fixed threshold for suppression of the faint background restoration errors caused by the intrinsic stochasticity of DDRMs. We pinpoint the window position that contains the most ``1'' pixels as the localized adversarial patch position (Step 5 in Figure \ref{fig:teaser}). Finally, DiffPAD masks the localized patch region and reconstructs the visual content in it by diffusion restoration conditioning on the surrounding unaltered region (Steps 6-7 in Figure \ref{fig:teaser}).

\section{Experiments}
\label{sec:experiments}

\subsection{Experimental setup}

This section introduces our main experimental setup. Other details are deferred to the supplementary materials. 

\vspace{0.1cm}

\shortsection{Datasets and networks} We evaluate DiffPAD by an image classification task on ImageNet~\cite{deng2009imagenet} and a facial recognition task on VGG Face~\cite{parkhi2015deep}. We employ up-to-date pre-trained classifiers ConvNeXt~\cite{liu2022convnet} and Swin-V2~\cite{liu2022swin}, which represent the most advanced architectures in convolution neural networks (CNNs) and vision transformers (ViTs), respectively. We use Inception-V3~\cite{szegedy2016rethinking} for adaptive attacks as in DIFFender~\cite{kang2023diffender} to compare their results with ours. We take advantage of the pre-trained diffusion model from~\cite{dhariwal2021diffusion}.

\vspace{0.1cm}

\shortsection{Attacks} We evaluate DiffPAD with three different patch attacks. AdvP~\cite{brown2017adversarial} is the standard localized perturbation with random position. LaVAN~\cite{karmon2018lavan} enhances the gradient updates of AdvP and delivers stronger attacks. GDPA~\cite{xiang2021gdpa} optimizes the patch’s position and pattern by an extra generative network. For AdvP and LaVAN, the number of attack iterations is set as $500$. For GDPA, the default $50$ epochs are employed. We also consider the white-box scenario, i.e., leveraging adaptive attack for comprehensive assessment. Since DiffPAD and selected baselines are in nature preprocessing mechanisms, we approximate obfuscated gradients via BPDA~\cite{athalye2018obfuscated}, assuming the output of the defense function equals the clean input. For BPDA-AdvP and BPDA-LaVAN, the number of attack iterations is set as $100$.

\vspace{0.1cm}

\shortsection{Baselines} We choose defense baselines that can serve as purification on input images, including smoothing-based defense JPG~\cite{dziugaite2016study}, segmentation-based defense SAC~\cite{liu2022segment}, entropy-based defense Jedi~\cite{tarchoun2023jedi}, and diffusion-based defense DiffPure~\cite{nie2022DiffPure} and DIFFender~\cite{kang2023diffender}. Among them, SAC, Jedi and DIFFender are specialized defenses against adversarial patches. Except for DIFFender which is not open-source, all other baselines are executed by their original implementation taking default parameter settings.

\vspace{0.1cm}

\shortsection{Evaluation metrics} The primary metrics for evaluating defenses are clean and robust accuracies under patch attacks. We evaluate the faithfulness of images post-defense, compared to clean images by Peak Signal-to-Noise Ratio (PSNR). In addition, we compute the mean Intersection over Union (mIoU) between the estimated patch region and the ground truth, which is an auxiliary metric to reflect the accuracy of our patch detection module. For all these metrics, a higher value indicates a better performance.

\subsection{Main results}

\begin{table}[t]
\caption{Comparisons of clean and robust accuracies (\%) on ImageNet with ConvNeXt across different patch defenses.}
\label{tab:convnext}
\vspace{-0.1cm}
\centering
\begin{tabular}{l|cccc}
\toprule
\diagbox{Defense}{Attack} & Clean & AdvP & LaVAN & GDPA \\ \midrule
w/o defense & 83.9 & 4.7 & 3.9 & 77.1                \\
JPG~\cite{dziugaite2016study} & 77.3 & 74.9 & 73.9 & 76.0                        \\
SAC~\cite{liu2022segment} & 80.6 & 79.8 & 80.0 & 79.4                        \\
Jedi~\cite{tarchoun2023jedi} & 82.2 & 80.3 & 80.8 & 80.1                        \\
DiffPure~\cite{nie2022DiffPure} & 76.4 & 74.3 & 74.5 & 75.6                    \\ \midrule
DiffPAD & \textbf{82.3} & \textbf{82.3} & \textbf{82.2} & \textbf{80.4} \\
\bottomrule
\end{tabular}
\end{table}

\begin{table}[t]
\caption{Comparisons of clean and robust accuracies (\%) on ImageNet with Swin-V2 across different patch defenses.}
\label{tab:swin}
\vspace{-0.1cm}
\centering
\begin{tabular}{l|cccc}
\toprule
\diagbox{Defense}{Attack} & Clean & AdvP & LaVAN & GDPA \\ \midrule
w/o defense & 83.4 & 0.5 & 0 & 75.7                     \\
JPG~\cite{dziugaite2016study} & 77.0 & 74.9 & 74.3 & 75.8                     \\
SAC~\cite{liu2022segment} & 81.5 & 81.1 & 80.8 & 79.6                     \\
Jedi~\cite{tarchoun2023jedi} & 81.3 & 79.5 & 79.4 & 79.0                        \\
DiffPure~\cite{nie2022DiffPure} & 76.9 & 75.7 & 75.3 & 76.3                      \\ \midrule
DiffPAD & \textbf{81.7} & \textbf{82.1} & \textbf{81.4} & \textbf{80.1} \\
\bottomrule
\end{tabular}
\end{table}

\begin{table}[t]
\caption{Comparisons of clean and robust accuracies (\%) under adaptive BPDA attacks with Inception-V3. Results of baselines are directly drawn from DIFFender~\cite{kang2023diffender}.}
\label{tab:adaptive}
\vspace{-0.1cm}
\centering
\begin{tabular}{l|ccc}
\toprule
\diagbox{Defense}{BPDA attack} & Clean & AdvP & LaVAN \\
\midrule
w/o defense & {100.0} & {0.0} & {8.2} \\
JPG~\cite{dziugaite2016study} & {48.8} & {0.4} & {15.2} \\
SAC~\cite{liu2022segment} & {92.8} & {84.2} & {65.2} \\
Jedi~\cite{tarchoun2023jedi} & {92.2} & {67.6} & {20.3} \\
DiffPure~\cite{nie2022DiffPure} & {65.2} & {10.5} & {15.2} \\
DIFFender~\cite{kang2023diffender} & {91.4} & {88.3} & {71.9} \\
\midrule
DiffPAD & \textbf{94.1} & \textbf{90.0} & \textbf{88.3} \\
\bottomrule
\end{tabular}
\end{table}

Table~\ref{tab:convnext} and Table~\ref{tab:swin} showcase the superior performance of DiffPAD, which outperforms all baselines on both clean and attacked images using AdvP, LaVAN and GDPA. While the accuracy gap between DiffPAD and the second-best baseline is marginal for clean data and GDPA attack, the improvement is significant for AdvP and LaVAN attacks. Notably, on ConvNeXt under AdvP attack, DiffPAD exceeds the second-ranked baseline by $2.0\%$. Generally, the clean or robust accuracy of JPG and DiffPure cannot surpass $78\%$, while SAC, Jedi, and DiffPAD are all higher than $79\%$. Evidently, global defenses fall short compared to specialized patch defenses under patch attacks.
Without defense, both ConvNeXt and Swin-V2 report highly accurate predictions on clean data. However, their performance will significantly drop under AdvP and LaVAN attacks. In particular, almost no images remain unscathed by AdvP or LaVAN attacks on Swin-V2, with accuracy plummeting to $0.5\%$ and $0\%$, respectively. After applying DiffPAD, the accuracy will be recovered to around 82\%, where the reduction compared to the clean accuracy is less than $2\%$ for both ConvNeXt and Swin-V2. Conversely, the GDPA attack results in a minor accuracy reduction, namely $6.8\%$ on ConvNext and $7.7\%$ on Swin-V2, yet defending against this attack proves to be more challenging. Neither DiffPAD nor selected baselines can bring the accuracy level back to above $81\%$. Given GDPA's inadequate attack performance on ConvNeXt and Swin-V2, we will mainly focus on DiffPAD under AdvP and LaVAN attacks in the following experiments.

An interesting observation is that Jedi performs better on ConvNeXt in all scenarios, ranking as the second-best defense, while SAC serves as the second-best on Swin-V2. This distinction is likely because the features extracted by ViTs are more globally entangled. In other words, when SAC blocks the visual content corresponding to the adversarial patches, ViTs still receive contextual information from other regions. In contrast, CNN-based architectures focus more on localized details, making them more susceptible to information loss. For CNNs, employing Jedi to disrupt the pattern of adversarial patches with their surrounding pixels is a better choice. This explains why ConvNeXt is more robust than Swin-V2 in the absence of defenses. Adversarial patches influence all semantic patches in the token operations of ViTs, while their impact on CNNs is less pronounced, because the small convolution kernels limit the effective propagation of localized information overall. 

\vspace{0.1cm}

\shortsection{Adaptive attacks} Table~\ref{tab:adaptive} presents the results of the BPDA adaptive attacks on Inception-V3. We observe that JPG and DiffPure significantly underperform, achieving less than $20\%$ accuracy under both AdvP and LaVAN attacks. This suggests that the gradients of such global, nuanced rectification methods are easier to be approximated than those of localized, drastic modification methods. Jedi exhibits stronger resilience against the AdvP-BPDA attack but fails to generalize to the LaVAN-BPDA attack. On the contrary, DiffPAD maintains consistent performance across different networks and attacks, whether adaptive or not. While SAC and DIFFender show commendable adversarial robustness in adaptive settings, they are still inferior to DiffPAD. For instance, DiffPAD outperforms DIFFender by $16.4\%$ under the LaVAN-BPDA attack, likely due to its more exact localization capabilities for adversarial patches.

\begin{figure*}[t]
    \centering
    \includegraphics[width=0.95\linewidth]{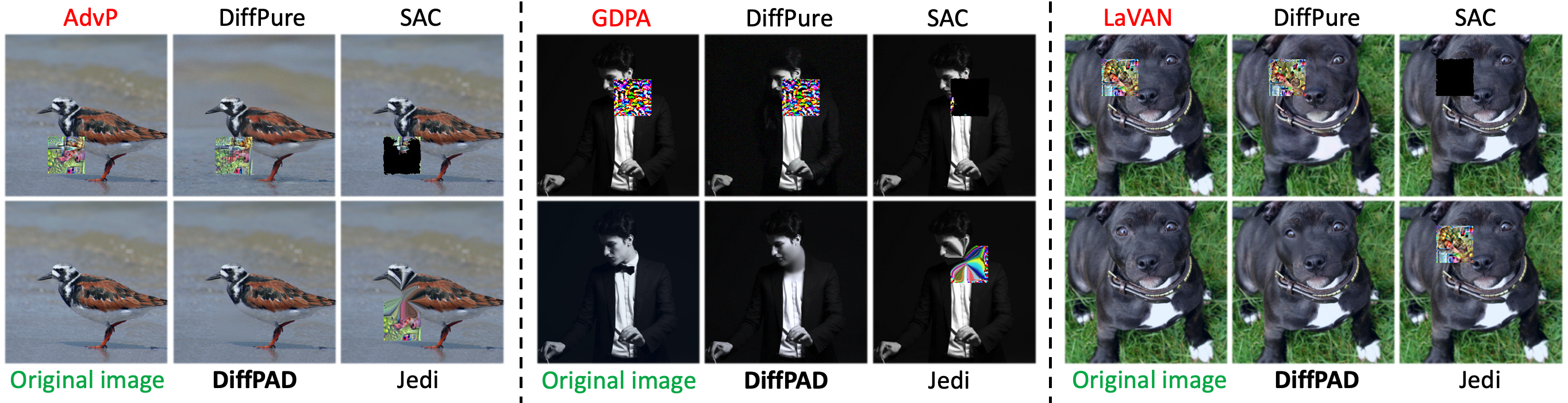}
    \vspace{-0.1cm}
    \caption{Illustration of three exampled visual effects on adversarial patches before and after applying different patch defenses. Note that it is difficult to find any traces of the adversarial patch from the images decontaminated by DiffPAD.}
    \label{fig:examples}
    \vspace{-0.1cm}
\end{figure*}

\begin{table}[t]
\caption{Effects of different modules in DifffPAD. PAD, INP and SVD stand for patch detection, inpainting restoration and singular value decomposition, respectively.}
\label{tab:ablation}
\vspace{-0.1cm}
\centering
\begin{tabular}{l|cc|cc}
\toprule
\multirow{2}{*}{\diagbox{Defense}{Attack} } & \multicolumn{2}{c|}{ConvNeXt (\%)} & \multicolumn{2}{c}{Swin-V2 (\%)} \\
\cmidrule{2-5}
& AdvP & LaVAN & AdvP & LaVAN \\
\midrule
DiffPAD w/o PAD & 70.7  & 69.5 & 72.9 & 71.7 \\
DiffPAD w/o INP & 80.7 & 80.1 & 81.2 & \textbf{82.1} \\
DiffPAD (SVD) & \textbf{82.5} & 82.0 & 81.8 & 81.9 \\
DiffPAD & 82.3 & \textbf{82.2} & \textbf{82.1} & 81.4 \\
\bottomrule
\end{tabular}
\end{table}

\begin{table}[t]
\caption{Patch localization precision in mIoU (\%) of DiffPAD with varying patch sizes and random positions.}
\label{tab:miou}
\vspace{-0.1cm}
\centering
\begin{tabular}{l|cc|cc}
\toprule
\multirow{2}{*}{\diagbox{Patch}{Attack} } & \multicolumn{2}{c|}{ConvNeXt} & \multicolumn{2}{c}{Swin-V2} \\
\cmidrule{2-5}
& AdvP & LaVAN & AdvP & LaVAN \\
\midrule
size 3\% & 82.27 & 83.34 & 80.42 & 80.57 \\
size 5\% & 85.10 & 83.40 & 86.66 & 86.31 \\
size 7\% & 83.35 & 83.44 & 86.52 & 86.69 \\ 
\bottomrule
\end{tabular}
\end{table}

\begin{figure}[t]
    \centering
    \includegraphics[width=0.95\linewidth]{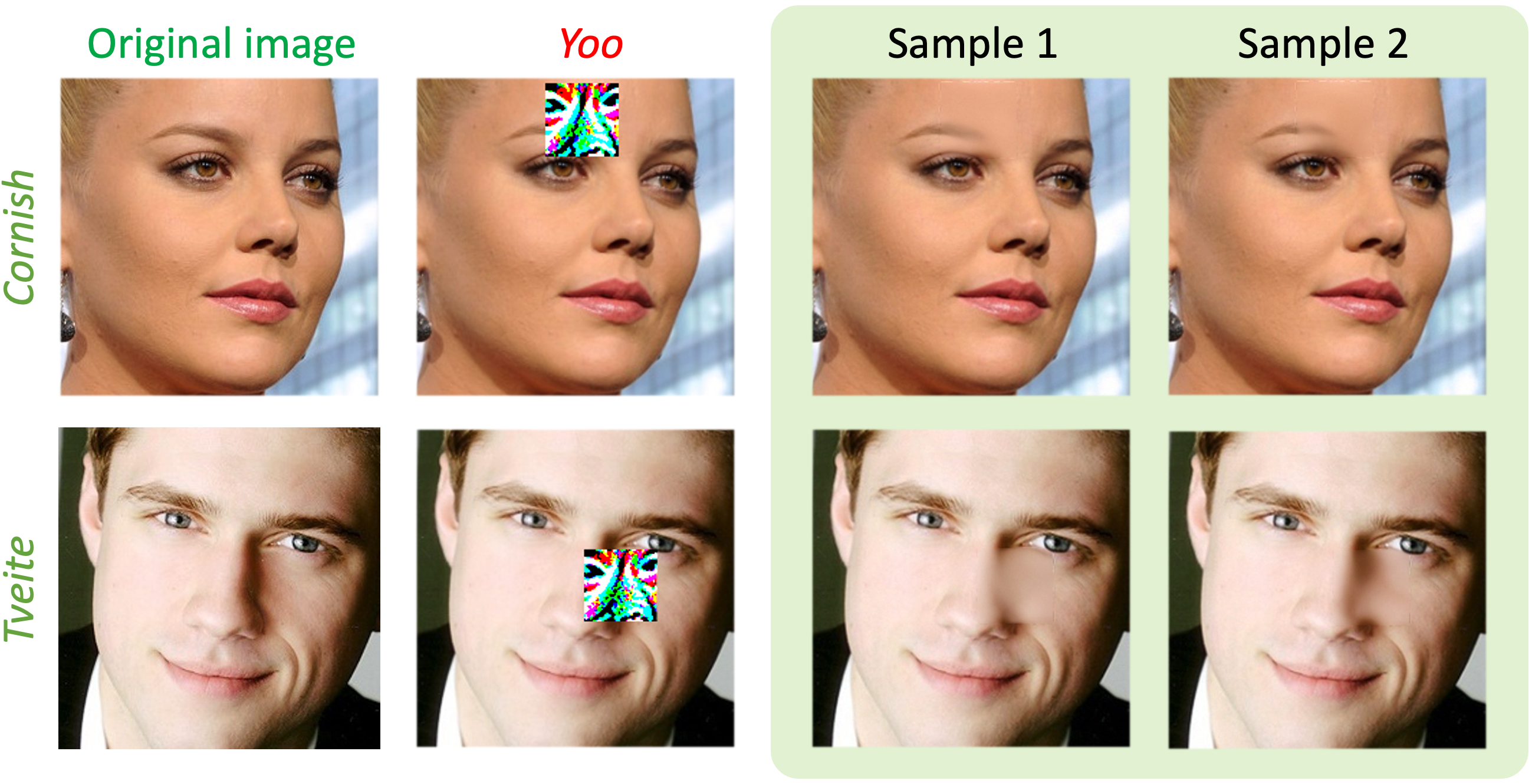}
    \vspace{-0.1cm}
    \caption{The performance of DiffPAD in facial recognition task on VGG Face. We run twice to attain two well-restored samples.}
    \label{fig:face}
    \vspace{-0.1cm}
\end{figure}

\vspace{0.1cm}

\shortsection{Ablation study} Table~\ref{tab:ablation} provides the results of an ablation study on each component of DiffPAD. Excluding patch detection means that we perform only resolution degradation-restoration on input images through a single diffusion generation. This yields even lower accuracy than the simplest JPG defense under the same attack conditions, as the resolution degeneration cannot be eliminated but rather mitigated. The blurring effect occurs globally, affecting both the structure of the adversarial patch as well as other visual details. As a result, Swin-V2 outperforms ConvNeXt by $2.2\%$ under either the AdvP or LaVAN attack, which is consistent with previous findings that CNNs are more sensitive to fine-grained visual features.

Including the patch detection module while excluding the inpainting restoration elevates DiffPAD to perform comparably with SAC under the same attacks. The final inpainting restoration step further promotes DiffPAD to a new SOTA in adversarial patch defense. Inpainting is necessary when a patch obscures critical semantics of the input images. It supplies meaningful visual content to the patch region, helping classifiers understand label semantics and preventing them from interpreting the mask as pertinent visual data. The increased stochasticity from more diffusion steps is also beneficial for resisting adaptive attacks. The last two rows in Table~\ref{tab:ablation} are outcomes of switching the closed-form solutions used in conditional sampling during the reverse diffusion process. The SVD solution~\cite{kawar2022denoising} has been explained at the end of Section~\ref{sec:preliminary}. The close accuracy levels under the same attacks imply that the influence of altering the conditional sampling strategy is small, suggesting the flexibility and stability of DiffPAD.

\vspace{0.1cm}

\shortsection{Varying patch attacks} Table \ref{tab:miou} demonstrates the patch localization precision under diverse patch attack conditions. It can be observed from the table that all the mIoU scores break through $80\%$, ensuring the generalizability of the patch detection module across varying patch sizes and random positions. This finding also validates the result of Theorem 1 with strong empirical evidence.
\begin{table}[t]
\caption{The faithfulness (PSNR) of images after various patch defenses with reference to the original clean images. Unit: dB.}
\label{tab:psnr}
\vspace{-0.1cm}
\centering
\begin{tabular}{l|cc|cc}
\toprule
\multirow{2}{*}{\diagbox{Defense}{Attack} } & \multicolumn{2}{c|}{ConvNeXt} & \multicolumn{2}{c}{Swin-V2} \\
\cmidrule{2-5}
& AdvP & LaVAN & AdvP & LaVAN \\
\midrule
w/o defense  & 21.25 & 20.70 & 22.26 & 22.01 \\
SAC~\cite{liu2022segment} & 19.63 & 19.57 & 19.42 & 19.42 \\
Jedi~\cite{tarchoun2023jedi} & 22.77 & 22.57 & 22.95 & 22.94 \\ \midrule
DiffPAD & \textbf{26.38} & \textbf{26.63} & \textbf{27.43} & \textbf{27.53} \\
\bottomrule
\end{tabular}
\end{table}

\subsection{Further analyses}

\shortsection{Visualizations} To examine the capability of removing patch remnants, we visualize the images returned by different methods. Figure~\ref{fig:examples} displays the visual effects on adversarial patches after applying baseline defenses and DiffPAD. DiffPure generally acts like a blurring function on the entire image, which is noticeable in the background of the first example. However, the pattern of adversarial patches cannot be washed out, confusing classifiers when the patch features compound with label-related features. SAC, while seldom misidentifying non-patch regions as adversarial, often fails to completely screen out all adversarial pixels, as shown in the first example. An apparent issue with Jedi is that it crushes estimated patch regions with distortions, injecting disruptive cues that interfere with recognition, especially for the second example. Jedi also omits the adversarial patch in the third example. In contrast, examples of DiffPAD conceal any patch remnants, with both naturalistic and meaningful visual content. Table~\ref{tab:psnr} lists that all PSNR values for DiffPAD are above 26 dB under all attack conditions, confirming the remarkable fidelity of the decontaminated adversarial examples referring to their original states. Additional visualization results in the supplementary materials also show our method has a negligible effect on clean images, ensuring low false positive rates of patch detection.

\vspace{0.1cm}

\shortsection{Transferability to facial recognition} To assess the transferability of DiffPAD, we apply the same hyperparameters selected for ImageNet experiments to the VGG Face in a facial recognition task. We also choose a new network--VGG16~\cite{Karen2015vgg} pre-trained by \cite{Wu2020Defending}, to test DiffPAD in a completely different task domain. We let GDPA attack the face images given its high success rate on VGG architectures~\cite{xiang2021gdpa}. By substituting the weights of the diffusion model with those pre-trained on the FFHQ~\cite{karras2019style} dataset as in~\cite{choi2021ilvr}, we obtain perfect facial restoration results, shown in Figure~\ref{fig:face}. VGG16 then regains the correct perdition with high confidence. The different samples convey the same success, manifesting the strong defensive capabilities of DiffPAD across various task domains.

\vspace{0.1cm}

\shortsection{Limitations and future works} Although our experiments illustrate the broad applicability of DiffPAD across various patch attack conditions, its practicability in real-time applications is constrained by the computational cost. It is worth noting that we optimize the efficiency of DiffPAD to execute only two rounds of diffusion generation per image, whereas DIFFender requires at least four rounds. Given the inherent complexity of diffusion models, the sampling process iterates multiple neural function evaluations (NFEs) at each time step, slowing down DiffPAD’s inference speed compared to SAC and Jedi. Another limitation of DiffPAD is that adversarial patches are assumed to be or can be enclosed by a square. This assumption might not hold for more irregularly shaped patches such as adversarial eyeglasses~\cite{sharif2019general}. Addressing these issues will be crucial for broadening the flexibility of DiffPAD in physical environments where such attacks may occur. Enhancing DiffPAD to overcome these limitations will increase its versatility as a tool for combating all adversarial patch attacks.

\section{Conclusion}
\label{sec:conclusion}

In this paper, we propose DiffPAD, a pretrained diffusion-based adversarial patch defense. DiffPAD first guide the conditional diffusion to restore the downsampled input images, where the diffusion restoration error is correlated to the patch size, informing a dynamic binarization and sliding window method for precise patch localization. DiffPAD then replaces the super-resolution with an inpainting solution to fill masked areas. Tested extensively across different attacks, patch sizes, target models, datasets, and task domains, DiffPAD boosts adversarial robustness and sustains naturalistic integrity of images, achieving SOTA performance without text guidance or fine-tuning.


\section*{Acknowledgment}
This work was in part financially supported by the Digital Futures. It was also part of the Swedish Wireless Innovation Network (SweWIN) approved by the Swedish Innovation Agency (VINNOVA). The computations were enabled by the resources provided by the National Academic Infrastructure for Supercomputing in Sweden (NAISS), partially funded by the Swedish Research Council.

{\small
\bibliographystyle{ieee_fullname}
\bibliography{egbib}
}
\end{document}


\title{Supplementary Materials for DiffPAD: Denoising Diffusion-based \\ Adversarial Patch Decontamination}

\author{Jia Fu$^{1,2}$\quad Xiao Zhang$^3$\quad Sepideh Pashami$^{1,4}$\quad Fatemeh Rahimian$^1$\quad Anders Holst$^{1,2}$\quad\\
$^1$RISE Research Institutes of Sweden\quad $^2$KTH Royal Institute of Technology\\
$^3$CISPA Helmholtz Center for Information Security\quad $^4$Halmstad University\\
{\tt\small \{jia.fu, sepideh.pashami, fatemeh.rahimian, anders.holst\}@ri.se\quad xiao.zhang@cispa.de}
}
\maketitle


\section{Additional discussions on related work}

In this section, we provide more detailed discussions of related works on adversarial patch attacks and diffusion-based adversarial defenses.


\subsection{Adversarial patch attacks} 

Since Szegedy~\textit{et al.}~\cite{szegedy2014intriguing} revealed the adversarial vulnerabilities of neural networks, where normal inputs crafted with imperceptible perturbations can induce erroneous predictions, numerous attack algorithms~\cite{goodfellow2015explaining, 7958570, madry2018towards} have been proposed to study the model behavior in the presence of adversarial examples. However, most existing works focused on global attacks defined by some $\ell_p$-norm, thereby not directly applicable to threatening real-world systems. Brown~\textit{et al.}~\cite{brown2017adversarial} first introduced the concept of adversarial patches, where the adversary is only allowed to manipulate a small region of an image to launch the evasion attack. Subsequently, LaVAN~\cite{karmon2018lavan} enhanced the design of the loss function, enabling the adversarial patch to cover only $2\%$ of the given image. Meanwhile, GDPA~\cite{xiang2021gdpa} improved the attack strategy by adversarially refining the patch's location rather than positioning it randomly. These research efforts lay the foundation for realizing adversarial patches in the physical world. For example, an adversarial patch printed on a T-shirt \cite{xu2020adversarial} can succeed in evading human detectors, while Wei~\textit{et al.}~\cite{wei2022adversarial} proposed adversarial stickers, which feature meaningful patterns and achieve good performance in both digital and physical realms.


\subsection{Diffusion-based adversarial defenses}
We further discuss the limitations of existing diffusion-based adversarial defenses, including DiffPure and DIFFender. DiffPure~\cite{nie2022DiffPure} has proved that forward diffusion disrupts the distribution of both clean data and adversarial perturbations. During the reverse diffusion process, clean data can be stochastically recovered, while adversarial effects are progressively eliminated. This process can be executed using the standard DDPM framework. Necessarily, to preserve the label semantics of the image, DiffPure halts the diffusion at a specific timestep $t^*\in(0, T)$ then commences the reverse diffusion from $\bm{x}_{t^*}$ back to $\bm{x}_0$. DIFFender~\cite{kang2023diffender} identified a critical limitation of DiffPure in adversarial patch defense. DiffPure struggles to completely remove the adversarial patch, which requires a larger $t^*$, whereas a smaller $t^*$ is essential for maintaining image semantics. Alternatively, DIFFender retains image semantics with the aid of additional prompts and fine-tunes a text-guided diffusion model for patch localization and restoration. However, prompt learning introduces new challenges, as well as limited prior contained within the text prompts renders DIFFender less efficient, necessitating the generation of at least three samples per image to ensure robust patch localization.

\section{Proof of Theorem 1}

For the sake of completeness, we provide detailed proof of our main theoretical result presented in Section 4.2. Our proof technique mainly follows from the proof of Theorem 3.2 in \cite{nie2022DiffPure}. Below, we first restate the problem statement of Theorem 1 that we are going to prove.

\begin{theorem}
\label{thm:connection}
Assume $\|\bm\epsilon_\theta\left(\bm{x}_t\right)\|\leq C_\epsilon\sqrt{1-\bar{\alpha}_t}$ and let $\gamma:=\int_0^{T} \beta_t \mathrm{d} t$.  With probability at least $1-\xi$, the $\ell_2$ distance between the diffusion-purified image $\hat{\bm{x}^a}$ with adversarial patch and the corresponding clean image $\bm{x}^c$ satisfies:
\begin{equation}
\tag{12}
\|\hat{\bm{x}^a}-\bm{x}^c\| \leq \varepsilon\left|\mathbf{A}\right|+\gamma C_\epsilon+\sqrt{e^{\gamma}-1} \cdot C_\xi,
\end{equation}
where $\varepsilon$ is the $\ell_2$-norm bound of the patch, $C_\xi:=\sqrt{2 d+4 \sqrt{d \log \frac{1}{\xi}}+4 \log \frac{1}{\xi}}$, and $d$ is the input dimension.
\end{theorem}

\noindent \textbf{Proof}: For variance preserving SDE, given the adversarial example $\bm{x}^a$ defined in Equation 8, after the forward diffusion process, we have
\begin{equation}
\tag{15}
\bm{x}_T=\sqrt{\alpha_T} \cdot \bm{x}^a+\sqrt{1-\alpha_T} \cdot \bm{\epsilon}',
\end{equation}
where $\alpha_T=e^{-\int_0^T \beta_t \mathrm{d} t}$ and $\boldsymbol{\epsilon}' \sim \mathcal{N}\left(\mathbf{0}, \mathbf{I}_d\right)$. As diffusion-restored adversarial example $\hat{\bm{x}^a}$ does not have a closed-form solution, we apply an SDE solver with the Euler–Maruyama discretization, where the drift and diffusion coefficients of the reverse-time SDE are given by:
\begin{equation}
\tag{16}
\begin{aligned}
\boldsymbol{f}_{\mathrm{rev}}(\boldsymbol{x}, t) & :=-\frac{1}{2} \beta_t\left[\boldsymbol{x}+2 \boldsymbol{s}_\theta(\boldsymbol{x}_t)\right], \\
g_{\mathrm{rev}}(t) & :=\sqrt{\beta_t},
\end{aligned}
\end{equation}
where $\boldsymbol{s}_\theta(\boldsymbol{x}_t)$ denotes the score function. The $\ell_2$ distance between $\hat{\bm{x}^a}$ and the corresponding clean data $\bm{x}^c$ can be bounded as:
\begin{equation}
\tag{17}
\begin{aligned}
& \|\hat{\bm{x}^a}-\boldsymbol{x}^c\| =\left\|\boldsymbol{x}_T+\left(\hat{\bm{x}^a}-\boldsymbol{x}_T\right)-\boldsymbol{x}^c\right\| \\
& =\|\boldsymbol{x}_T+\int_T^0-\frac{1}{2} \beta_t\left[\boldsymbol{x}+2 \boldsymbol{s}_\theta(\boldsymbol{x}_t)\right] \mathrm{d} t+\int_T^0 \sqrt{\beta_t} \mathrm{d} \boldsymbol{w}-\boldsymbol{x}^c\| \\
& \leq\|\underbrace{ \boldsymbol{x}_T+\int_T^0-\frac{1}{2} \beta_t \boldsymbol{x} \mathrm{d} t+\int_T^0 \sqrt{\beta_t} \mathrm{d} \boldsymbol{w}}_{\text {Integration of linear SDE }}-\boldsymbol{x}^c\| \\
&\qquad\qquad\qquad\qquad\qquad\qquad+\|\int_T^0-\beta_t \boldsymbol{s}_\theta(\boldsymbol{x}_t) \mathrm{d} t\|,
\end{aligned}  
\end{equation}
where the second equation is obtained by using the integration of the reverse-time SDE, and the last line is derived by separating the integration of the linear SDE from non-linear SDE involving $\boldsymbol{s}_\theta(\boldsymbol{x}_t)$ through the triangle inequality.

Notice that the above linear SDE is a time-varying Ornstein–Uhlenbeck process, where the time increment inversely starts from $T$ to $0$ with the initial value $\bm{x}_T$. Denote its solution by $\bm{x}'$ that follows a Gaussian distribution, the mean $\bm{\mu}_0$ and covariance matrix $\bm{\Sigma}_0$ of $\bm{x}'$ will be the solutions of the following two differential equations:
\begin{equation}
\tag{18}
\begin{aligned}
\frac{\mathrm{d} \boldsymbol{\mu}}{\mathrm{d} t} &=-\frac{1}{2} \beta_t \boldsymbol{\mu}, \\
\frac{\mathrm{d} \boldsymbol{\Sigma}}{\mathrm{d} t} &=-\beta_t \boldsymbol{\Sigma}+\beta_t \mathbf{I}_d,
\end{aligned}
\end{equation}
with the initial conditions $\bm{\mu}_T=\bm{x}_T$ and $\bm{\Sigma}_T=\bm{0}$. By solving these two differential equations, we have $\bm{x}'\sim \mathcal{N}\left(e^{\frac{\gamma}{2}} \boldsymbol{x}_T,\left(e^{\gamma}-1\right) \mathbf{I}_{d}\right)$ that is conditioned on $\bm{x}_T$, where $\gamma:=\int_0^{T} \beta_t \mathrm{d} t$.
Taking the advantage of reparameterization trick, we obtain
\begin{equation}
\tag{19}
\begin{aligned}
&\boldsymbol{x}^{\prime}-\boldsymbol{x}^c \\ &=e^{\frac{\gamma}{2}} \boldsymbol{x}_T+\sqrt{e^{\gamma}-1} \cdot \boldsymbol{\epsilon}''-\boldsymbol{x}^c \\
& =e^{\frac{\gamma}{2}}\left(e^{-\frac{\gamma}{2}}\boldsymbol{x}^a+\sqrt{1-e^{-\gamma}} \cdot \boldsymbol{\epsilon}'\right)+\sqrt{e^{\gamma}-1} \cdot \boldsymbol{\epsilon}''-\boldsymbol{x}^c \\
& =\sqrt{e^{\gamma}-1}\cdot\left(\boldsymbol{\epsilon}'+\boldsymbol{\epsilon}''\right)+\boldsymbol{x}^a-\boldsymbol{x}^c,
\end{aligned}
\end{equation}
where the second equation follows by substituting Equation 15. Since $\boldsymbol{\epsilon}'' \sim \mathcal{N}\left(\mathbf{0}, \mathbf{I}_d\right)$ and $\boldsymbol{\epsilon}'\perp\boldsymbol{\epsilon}''$, the first term of the last line in Equation 19 can be combined as a zero-mean Normal variable with variance $2\left(e^{\gamma}-1\right)$. 

We know the connection between the score function and the noise prediction $\boldsymbol{\epsilon}_\theta(\boldsymbol{x}_t)$ in DDPM can be formulated as:
\begin{equation}
\tag{20}
\boldsymbol{s}_\theta(\boldsymbol{x}_t) =-\frac{\boldsymbol{\epsilon}_\theta(\boldsymbol{x}_t)}{\sqrt{1-\bar{\alpha}_t}}.
\end{equation}
Assuming that the $\ell_2$-norm of $\boldsymbol{\epsilon}_\theta(\boldsymbol{x}_t)$ is upper-bounded by $C_\epsilon\sqrt{1-\bar{\alpha}_t}$. In other words, we assume that the $\ell_2$-norm of $\boldsymbol{s}_\theta(\boldsymbol{x}_t)$ is upper-bounded by constant $C_\epsilon$. Hence, 
\begin{equation}
\tag{21}
\begin{aligned}
\|\hat{\bm{x}^a}-\bm{x}^c\| & \leq\|\sqrt{2\left(e^{\gamma}-1\right)}\cdot\boldsymbol{\epsilon}+\boldsymbol{x}^a-\boldsymbol{x}^c\|+\gamma C_\epsilon \\
& \leq\|\boldsymbol{x}^a-\boldsymbol{x}^c\|+\gamma C_\epsilon+\sqrt{2\left(e^{\gamma}-1\right)}\cdot\|\boldsymbol{\epsilon}\|,
\end{aligned}
\end{equation}
where $\boldsymbol{\epsilon} \sim \mathcal{N}\left(\mathbf{0}, \mathbf{I}_d\right)$. We denote the $\ell_2$-norm bound of the pixels in adversarial patch region as $\varepsilon$, since $\boldsymbol{x}^a-\boldsymbol{x}^c=\mathbf{A}\odot\left(\pmb{\delta}-\boldsymbol{x}^c\right)$, we can obtain $\|\boldsymbol{x}^a-\boldsymbol{x}^c\|\leq \varepsilon\left|\mathbf{A}\right|$, where $\left|\mathbf{A}\right|$ represents the pixel number, i.e., the size of adversarial patch. Furthermore, $\|\boldsymbol{\epsilon}\|^2 \sim \chi^2(d)$,  from the concentration inequality, we attain
\begin{equation}
\tag{22}
\operatorname{Pr}\left(\|\boldsymbol{\epsilon}\|^2 \geq d+2 \sqrt{d \sigma}+2 \sigma\right) \leq e^{-\sigma}.
\end{equation}
Let $e^{-\sigma}=\xi$, we get
\begin{equation}
\tag{23}
\operatorname{Pr}\left(\|\boldsymbol{\epsilon}\| \geq \sqrt{d+2 \sqrt{d \log \frac{1}{\xi}}+2 \log \frac{1}{\xi}}\right) \leq \xi.
\end{equation}
Finally, at least of the probability $1-\xi$, we have
\begin{equation}
\tag{24}
\|\hat{\bm{x}^a}-\bm{x}^c\| \leq \varepsilon\left|\mathbf{A}\right|+\gamma C_\epsilon+\sqrt{e^{\gamma}-1} \cdot C_\xi,
\end{equation}
where constant $C_\xi:=\sqrt{2 d+4 \sqrt{d \log \frac{1}{\xi}}+4 \log \frac{1}{\xi}}$, which completes the proof of Theorem 1.

\section{Experimental details}

\subsection{Hyperparameter setup}
All our experiments are conducted in Pytorch on four Nvidia A100 GPUs. We set $\mu = 0.066$ and $\nu = 14.90$ in Equation 14, which is determined using grid search. In practice, to reduce the redundant computations, the threshold $\tau'$ is fixed as $9$. We treat input images with diffusion restoration errors less than $62$ as clean images to prevent excess defense. We run $20$ NFEs for both super-resolution and inpainting restoration. Noise level $\sigma = 0.001$ and scaling factor $s = 4$ are hyperparameters in close-form solutions (Equation 10, 11). Additionally, we repeat three rounds of each experiment related to DiffPAD and report averaged statistics, due to the stochasticity of diffusion processes. In the evaluation phase, we adopt the same subset of the original ImageNet validation set as~\cite{pan2020dgp}, which contains $1000$ images covering all categories. For a fair comparison with DIFFender, we randomly choose $512$ images from this subset which can be correctly classified before the attacks.

\begin{figure}
    \centering
    \includegraphics[width=0.95\linewidth]{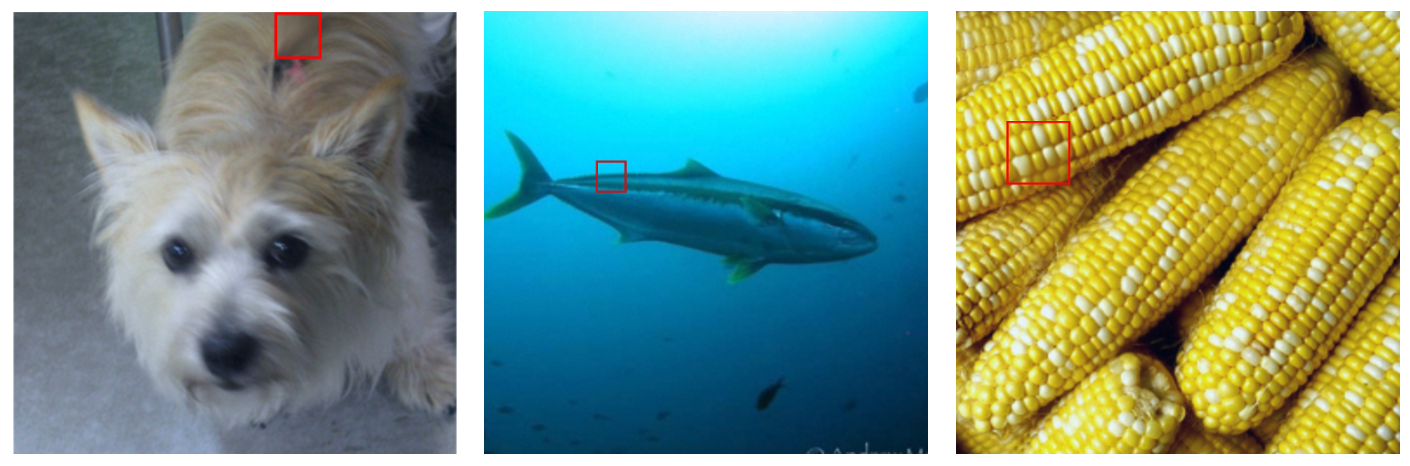}
    \vspace{-0.1cm}
    \caption{Examples of clean images where DiffPAD spuriously detects an adversarial patch of small size (marked by the red box).}
    \label{fig:fake}
\end{figure}

\subsection{False positive of patch detection}
Figure~\ref{fig:fake} visualizes how clean images appear when processed with DiffPAD. We can see that the estimated patches are quite small. The inpainting is competent in recovering an image almost identical to its original version, thereby avoiding excessive defense and ensuring the recognition performance remains unaffected on the clean dataset. This is also confirmed by the clean accuracies of DiffPAD, which is always the highest compared to the other defenses.

\subsection{Computational complexity}
For each image resized to $256\times256$, SAC~\cite{liu2022segment} costs $0.27$s, Jedi~\cite{tarchoun2023jedi} costs $0.32$s, DiffPAD costs $2.45$s, and DiffPure costs $8.59$s, on average.

\section{Generalizability to global attacks}
Although DiffPAD targets localized patch attacks, the proposed diffusion-based resolution degradation-restoration mechanism can serve as a handy tool to mitigate $\ell_p$-norm bounded perturbations. Table~\ref{tab:global} compares the robust accuracies of DiffPAD with other baselines used in the main paper against FGSM~\cite{goodfellow2015explaining}, PGD~\cite{madry2018towards}, and C\&W~\cite{7958570} attacks. The trivial image transformation and other patch defenses demonstrate limited effectiveness, far less than the SOTA model DiffPure in such attack settings. However, DiffPAD (40 NFEs) is second only to DiffPure and achieves $80\%$ of its performance, taking only $30\%$ of its runtime.

\begin{table}[t]
\caption{Comparisons of robust accuracies (\%) against global attacks on ImageNet with Inception-V3. The best (blue) and second-best (red) results are highlighted. PAD stands for patch detection.}
\label{tab:global}
\vspace{-0.1cm}
\centering
\begin{tabular}{l|cccc}
\toprule
\diagbox{Defense}{Attack} & FGSM & PGD & C\&W \\ \midrule
w/o defense & 14.3 & 0.2 & 0.1                     \\
JPG & 27.6 & 10.6 & 34.9                     \\
SAC & 19.6 & 2.8 & 4.0                     \\
Jedi & 25.9 & 5.6 & 22.5                     \\
DiffPure & \textcolor{blue}{64.4} & \textcolor{blue}{64.6} & \textcolor{blue}{65.8}                       \\ \midrule
DiffPAD w/o PAD & \textcolor{red}{50.3} & \textcolor{red}{51.1} & \textcolor{red}{53.3} \\
\bottomrule
\end{tabular}
\end{table}

{\small
\bibliographystyle{ieee_fullname}
\bibliography{egbib}
}